\documentclass{article}

\usepackage[final]{neurips_2025}

\usepackage[utf8]{inputenc} 
\usepackage[T1]{fontenc}    
\usepackage{hyperref}       
\usepackage{url}            
\usepackage{booktabs}       
\usepackage{amsfonts}       
\usepackage{nicefrac}       
\usepackage{microtype}      
\usepackage{xcolor}         
\usepackage{graphicx}
\usepackage{amsmath}
\usepackage{xcolor}

\definecolor{turquoise}{rgb}{0.25,0.88,0.82}

\title{ Causality Is Key to Understand and Balance Multiple Goals in Trustworthy ML and Foundation Models}

\author{%
\parbox{\linewidth}{\centering
  Ruta Binkyte\thanks{Equal contribution. Correspondence: \texttt{ruta.binkyte@gmail.com}.}\textsuperscript{1}
  \quad Ivaxi Sheth\footnotemark[1]\textsuperscript{1}
  \quad Zhijing Jin\textsuperscript{2,4,5}
  \quad Mohammad Havaei\textsuperscript{3}
  \quad Bernhard Sch\"{o}lkopf\textsuperscript{2}
  \quad Mario Fritz\textsuperscript{1}
  \\[0.8em]  
  \textsuperscript{1}CISPA Helmholtz Center for Information Security \quad
  \textsuperscript{2}Max Planck Institute for Intelligent Systems, T\"{u}bingen \\
  \textsuperscript{3}Google Research \quad
  \textsuperscript{4}ETH Z\"{u}rich \quad
  \textsuperscript{5}University of Toronto
}}

\begin{document}

\maketitle

\begin{abstract}
Ensuring trustworthiness in machine learning (ML) systems is crucial as they become increasingly embedded in high-stakes domains. This paper advocates for integrating causal methods into machine learning to navigate the trade-offs among key principles of trustworthy ML, including fairness, privacy, robustness, accuracy, and explainability. While these objectives should ideally be satisfied simultaneously, they are often addressed in isolation, leading to conflicts and suboptimal solutions. Drawing on existing applications of causality in ML that successfully align goals such as fairness and accuracy or privacy and robustness, \textbf{this position paper argues that the causal approach is essential for balancing multiple competing objectives in both trustworthy ML and foundation models.}
Beyond highlighting the potential to resolve trade-offs in trustworhy ML, we examine how causality can be practically integrated into ML and foundation models. Finally, we discuss the challenges, limitations, and opportunities in adopting causal frameworks, paving the way for more accountable and ethically sound ML systems.
\end{abstract}

\section{Introduction}


In recent years, machine learning (ML) has made remarkable strides, driving breakthroughs in natural language processing, computer vision, and decision-making systems~\cite{jia2024decision,achiam2023gpt}. These advancements have led to widespread adoption across diverse domains, including healthcare~\cite{singhal2025toward}, finance~\cite{lee2024survey}, education~\cite{team2024learnlm}, and social media~\cite{bashiri2024transformative}, where ML models now play a crucial role in diagnostics, algorithmic trading, personalized learning, and content recommendation.

Given their soaring influence, ensuring ethical and trustworthy ML systems has become a global priority~\cite{lewis2020global}. Many international regulations and frameworks \cite{EU_AI_Act_2021,OECD_AI_Principles_2019,G20_AI_Principles_2019,Singapore_AI_Governance_2020}
seek to establish guidelines for fairness, explainability, robustness, and privacy protection. 

Various organizations provide slightly different definitions and components of trustworthy ML. For the scope of our paper, we focus on five core dimensions that are both widely recognized and directly relevant to causal reasoning: fairness, privacy, robustness, explainability, and accuracy. We will introduce these dimensions and highlight their trade-offs and intersections below.
\begin{figure*}
    \centering
    \includegraphics[width= 0.8\linewidth]{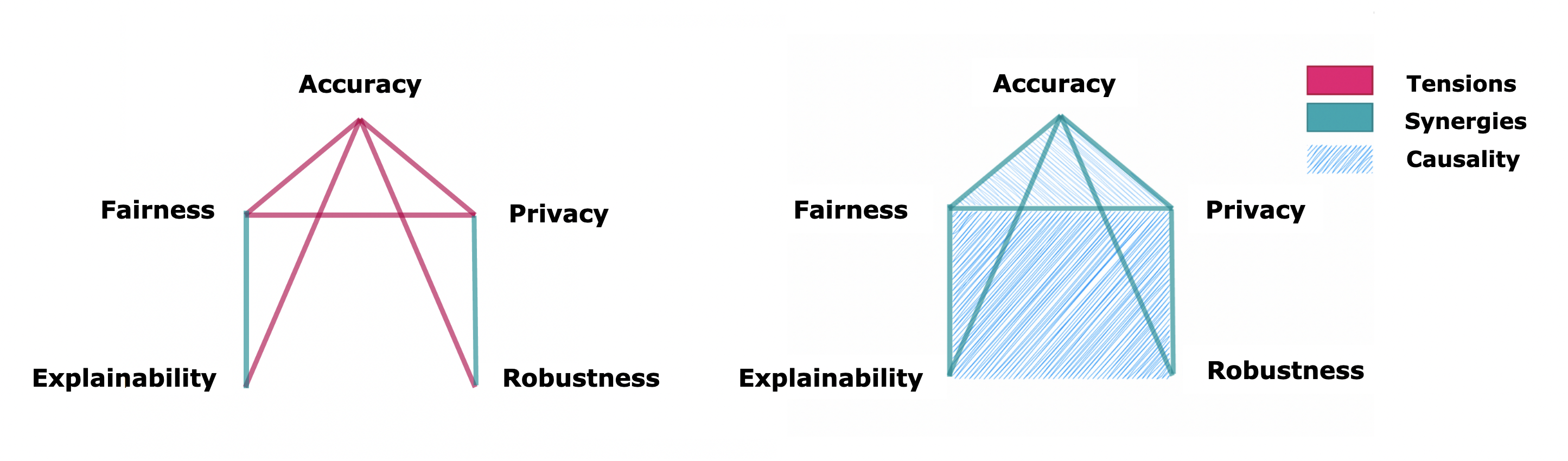}
    \vspace{-1em}
    \caption{While trustworthy ML involves inherent trade-offs between its key components, causality can help mitigate these tensions and enhance synergies.}
    \label{fig:tradeoffs}
\end{figure*}

Trustworthy ML must balance competing objectives like accuracy, fairness, robustness, privacy, and explainability. While trade-offs—such as between accuracy and fairness or privacy—are often unavoidable, some goals can be mutually reinforcing (e.g., explainability aiding fairness, privacy supporting robustness)~\cite{dwork2009differential,hopkins2022robustness}.

\textbf{Fairness.} Fairness in ML refers to the principle that systems should make unbiased decisions that do not discriminate against individuals or groups based on sensitive attributes such as race, gender, or socioeconomic status. ML systems have been shown to rely heavily on biased data, amplifying existing biases and leading to unequal outcomes~\cite{COMPAS}. These systems often exhibit reduced accuracy for minority or underrepresented groups, further exacerbating disparities~\cite{buolamwini2018gender}. Given the speed and scale of ML-enabled decisions, ensuring fairness is essential to prevent perpetuating and exacerbating societal inequalities at an unprecedented scale.

\textbf{Privacy.} Privacy in ML emphasizes the protection of individuals' sensitive and personal data. It has been shown that even after removing identifiers such as names, information can still leak, and individuals can be reidentified through indirect attributes and data triangulation\cite{sweeney2000simple,narayanan2008robust,ohm2010broken,dwork2006differential}. Additionally, sensitive information can be reconstructed from gradients during model training if data is not handled privately~\cite{zhu2019deep,geiping2020inverting,aono2017privacy,fredrikson2015model}. Privacy is crucial for ensuring compliance with data protection laws and safeguarding human rights. It also fosters trust for individuals to be more willing to contribute their data for model training if their safety and privacy are ensured.

\textbf{Robustness.}
Robustness refers to the system's ability to perform reliably under varying conditions, including adversarial attacks, noisy inputs, or distributional shifts. For example, models often underperform when faced with distribution shifts, such as changes in data characteristics between training and deployment environments~\cite{hendrycks2019benchmarking,recht2019imagenet,ovadia2019can}. Additionally, human-undetectable noise added to images can cause models to make incorrect predictions, highlighting their vulnerability~\cite{szegedy2014intriguing, goodfellow2015explaining}. Robustness is critical to ensuring the safety and reliability of AI systems, particularly in high-stakes applications such as healthcare and autonomous driving.

\textbf{Explainability.} Explainability refers to the ability of AI systems to provide clear and understandable reasoning behind their decisions or predictions. Deep neural networks (DNNs), often referred to as ``black boxes,'' are inherently complex and difficult to interpret, making them hard to audit and assess for fairness or correctness~\cite{lipton2018mythos, doshi2017towards,rudin2019stop}. Explainability is closely tied to accountability, as it enables stakeholders to evaluate and challenge AI outputs when necessary. Furthermore, regulations such as the GDPR emphasize the ``right to explanation,'' which requires that individuals be informed about and understand how automated decisions affecting them are made \cite{GDPR_site}.

\textbf{Trade-offs and Intersections.}
The trustworthy ML landscape involves complex trade-offs and interdependencies between key objectives such as fairness, privacy, accuracy, robustness, and explainability (~\autoref{fig:tradeoffs}). Improving one aspect often comes at the expense of another, such as the trade-off between \textbf{privacy} and accuracy in differential privacy, where noise added to protect data reduces model accuracy~\cite{xu2017privacy,carvalho2023towards}. Similarly, achieving \textbf{fairness} frequently requires sacrificing predictive performance or resolving conflicts between competing fairness notions, such as demographic parity and equalized odds~\cite{friedler2021possibility,kim2020fact}. Trade-offs also arise in \textbf{explainability} and accuracy, as complex models like DNNs excel in performance but lack interpretability. Meanwhile, the relationship between fairness and privacy is nuanced, with evidence showing they can either conflict, as noise may lead to disparate outcomes, or complement each other by reducing bias~\cite{pujol2020fair,dwork2011fairness}.

\textbf{Causality.} One of the most influential causal frameworks is Pearl's structural causal models (SCMs), which provide a systematic approach to reasoning about causality and integrating it into machine learning~\cite{pearl_causality_2009}. This framework defines causality as the relationship between the variables where a change in one variable (\textit{the cause}) directly leads to a change in another variable (\textit{the effect}). It establishes a directional and often mechanistic link, distinguishing relationships arising from mere correlations.

A key component of Pearl’s framework is the use of directed acyclic graphs (DAGs) and do-calculus, which offer a structured representation of causal dependencies and a formal method for performing causal inference. While important for causal reasoning, recent advances have proposed approximate methods for applying causality with partial causal graphs or discover implicit causal graphs from the data, making causal approaches more applicable in practice~\cite{kaddour2022causal,burgess2018understanding, kim2021counterfactual,russell2017worlds,ehyaeibridging,zuo2022counterfactual}.


Unlike correlation-based approaches, causality provides a framework for disentangling the underlying mechanisms that drive observed phenomena, offering a deeper interpretation of data. 
Causal frameworks have been successfully applied to audit and mitigate fairness~\cite{kim2021counterfactual,kilbertus2017avoiding,loftus18} 
and to improve robustness~\cite{scholkopf2022causality}. The research about the connection between causality and privacy is still very limited, but some emerging studies show potential for applications~\cite{tschantz2020sok, tople2020alleviating, binkyte2024causal}. Finally, explainability is one of the core features of causality and comes pre-packaged with the causal framework. Despite the promising applications of causality for individual requirements of trustworthy AI, the potential to use causality to reconcile multiple requirements of trustworthy ML remains largely under-explored.

\textbf{Position.} Despite significant advancements in research on individual dimensions of trustworthy ML such as fairness, privacy, and explainability—there is a notable lack of efforts to integrate these dimensions into a cohesive and unified framework. Each ethical principle addresses distinct challenges, yet their interplay often involves intricate trade-offs, particularly concerning model performance metrics such as accuracy. For example, mitigating fairness-related biases may require adjustments that compromise predictive precision, while enhancing explainability can impose constraints on model complexity. We argue that systematically addressing these trade-offs is a critical step toward developing AI systems that are both ethically sound and operationally efficient. 
While causality has been applied to address individual challenges such as fairness or interpretability, its potential to address the intersection of these challenges has largely been overlooked.
\textbf{In this position paper, we argue that integrating causality into ML and foundation models offers a way to balance
multiple competing objectives of trustworthy AI. }
We base our position on a review of how causal machine learning has advanced various aspects of trustworthy ML. From this, we identify key mechanisms behind objective trade-offs and show how causal methods can help resolve them. We then explore efforts to integrate causality into foundation models, highlighting unique challenges and outlining future directions for applying causal reasoning in this new context.



The structure of our paper is as follows. Section~\ref{sec:CausaltrustML} discusses how causality can reconcile multiple dimensions of trustworthy ML and explores how it can be integrated. Section~\ref{sec:caualfoundation} discusses how FMs amplify existing ML trade-offs and introduce new challenges and propose causal approach to overcoming these issues as well as strategies for integrating causal reasoning into FMs.  Section~\ref{sec:challenges} covers limitations and  possible solutions, and Section~\ref{sec:Alternative} includes alternative views.  Finally, Section~\ref{sec:conclusion} suggests actionable steps for advancing causality in ML and FMs and proposes future research directions.

\section{Causality for  Trustworthy ML}
\label{sec:CausaltrustML}
Causal modeling provides a principled way to address these tensions by capturing the underlying data-generating processes. This section first explores how causal approaches can help reconcile key trade-offs, then outlines methods for integrating causality into machine learning.

\subsection{Causality for Trade-offs in Trustworthy ML}~\label{sec:trustML}
 In this section, we examine key trade-offs in trustworthy ML and illustrate how causal approaches can help reconcile these competing objectives.

 \textbf{Fairness vs. Accuracy.} 
Efforts to improve fairness in ML often reduce accuracy, as they may obscure predictive features or constrain outputs to meet fairness criteria~\cite{pinzon2022impossibility, cooper2021emergent, zliobaite2015relation, zhao2022inherent}. Many such trade-offs stem from focusing on correlations rather than underlying causes. Causal models offer a principled solution by isolating spurious pathways from valid predictive factors. For example, counterfactual fairness ensures predictions remain stable when sensitive attributes (e.g., race) are altered in a counterfactual world~\cite{kusner2017counterfactual}.
A compelling example comes from the COMPAS dataset, where Black defendants were more likely to be classified as high-risk for recidivism. While statistical methods attribute this to race, a causal model might reveal over-policing in Black neighborhoods as a confounder of recidivism predictions~\cite{neil2023racial}. Adjusting for such causal pathways allows for risk assessments that are both more accurate and fair~\cite{chiappa2019path, zafar2017fairness}. Causal frameworks thus mitigate bias without discarding valid predictive signals, effectively softening the fairness-accuracy trade-off.

\textbf{Conflicting Notions of Fairness.}
Fairness in ML is often constrained by conflicting definitions and measurement approaches. \cite{friedler2021possibility} highlight the fundamental tension between the ``what you see is what you get" and ``we are all equal" worldviews---where the former accepts disparities based on observed merit, while the latter seeks to correct historical inequalities. Causal graphs can crisply formulate different notions of fairness  \cite{nabi2018fair,3a2e333502f5035f0f78318c7722ec1a7b7265bb}, thus enabling feasible mitigation via path-specific causal effects \cite{avin2005identifiability}.

\cite{kim2020fact} formalize fairness conflicts using the fairness-confusion tensor, showing that notions like demographic parity and equalized odds impose incompatible constraints. The causal approach mitigates these conflicts by focusing on fairness as a property of causal pathways rather than statistical dependencies~\cite{rahmattalabi2022promises}. This allows for greater flexibility in aligning fairness interventions with real-world causal mechanisms, allowing better-informed choice of fairness metric.

\textbf{Explainability for Fairness.} A particularly powerful approach within causal explainability is counterfactual explanations, which help users understand model decisions by asking ``what if" questions. Counterfactual methods generate alternative scenarios where certain features are changed while keeping others constant, allowing for a direct assessment of how specific inputs influence predictions~\cite{wachter2017counterfactual, karimi2020model}. Counterfactual explanations are particularly useful for fairness auditing as they can help identify why certain groups are adversely affected and guide corrective measures.

\textbf{Privacy vs. Accuracy.} 
Differential privacy introduces noise—controlled by parameter $\epsilon$—to protect sensitive data, but this often reduces model accuracy~\cite{xu2017privacy,carvalho2023towards}. The trade-off between privacy and utility remains fundamentally unresolved. Causal models offer a more principled alternative by framing privacy breaches as causal effects—where adversaries infer sensitive attributes through observable pathways~\cite{tschantz2020sok}.

By aligning interventions with causal structure, models can obscure private information (e.g., race, sex) while preserving essential dependencies, minimizing accuracy loss. For instance, causal graphs can enforce realistic variable combinations (e.g., aligning age and education), preventing implausible data patterns and reducing the risk of leakage. This approach improves both privacy protection and model fidelity.

\textbf{Privacy vs. Fairness.}
Privacy mechanisms, such as noise addition, can disproportionately impact minority groups, leading to fairness concerns. Differentially Private Stochastic Gradient Descent (DP-SGD), for example, has been shown to degrade model accuracy more severely for underrepresented groups, exacerbating fairness disparities~\cite{bagdasaryan2019differential}. However,  Causal models can guide privacy interventions by ensuring that noise is applied in ways that do not disrupt fairness-critical relationships. For instance, a causal graph can reveal which features or pathways should be preserved to maintain fairness while protecting privacy.

\textbf{Privacy vs. Robustness.} 
Adding noise without considering the data structure or causal relationships can obscure meaningful patterns and introduce spurious correlations. This indiscriminate noise can make models less robust to unseen data, particularly under distribution shifts.

In contrast, causal models inherently emphasize invariant relationships—patterns that are stable across various data distributions. Noise that disrupts non-causal relationships or spurious correlations can further enhance the robustness of these models to shifts in data. Finally, some results show, that causal models provide stronger guarantees for adversarial robustness with lower epsilon in differential privacy, thus allowing for lesser negative impact on accuracy~\cite{tople2020alleviating}.

\textbf{Robustness vs. Accuracy.}
The trade-off between generalizability and accuracy is rooted in the observation that models trained to achieve high accuracy on a specific dataset often overfit to the peculiarities of that distribution. This overfitting compromises their ability to generalize to new, unseen distributions~\cite{scholkopf2022causality}.
On the contrary, causal models focus on invariant relationships that hold across different environments, making them robust to distribution shifts. This robustness enhances the model's ability to generalize to unseen data, improving accuracy in diverse settings. For example, causal representation learning disentangles stable causal factors, allowing the model to maintain performance when data distributions change. Moreover,~\cite{richens2024} prove that robust agents implicitly learn causal world models, further emphasizing the intrinsic interdependency between robustness and causality.

\textbf{Explainability vs. Accuracy.}
Many complex algorithms, such as deep neural networks (DNN) or random forest (RF), have impressive predictive power but provide ``black-box" solutions that are hard to question or evaluate~\cite{london2019artificial,van2021trading}.  Causal models offer inherently interpretable structures by quantifying the contribution of each input feature to the output, providing clear, human-understandable explanations. Causal recourse further enhances explainability by offering actionable recommendations for individuals affected by model decisions, helping them achieve a more favorable outcome~\cite{KarSchVal21}.




\textbf{Prediction Accuracy vs. Intervention Accuracy.} One of the key advantages of the causal framework is its ability to support not just prediction but also intervention~\cite{hernan2020causal, scholkopf2022causality}. While predictive models are sufficient in some domains, many high-stakes applications—such as healthcare, policy-making, personalized treatment, as well as addressing structural biases and fairness, require actionable interventions~\cite{kusner2018causal, kusner2019making}. In these settings, understanding causal relationships is essential, as the objective is not only to predict outcomes but also to influence them.

\subsection{Integrating Causality into ML}~\label{sec:causalML}
Integrating causality into ML allows models to move beyond surface-level patterns and capture the underlying mechanisms that generate data. This section introduces methods for incorporating causal knowledge into ML models, as well as techniques for using ML to discover causal relationships from data.

\textbf{Causally Constrained ML (CCML).} CCML refers to approaches that \textit{explicitly} incorporate causal relationships into model training or inference as constraints or guiding principles. Given a causal graph \( \mathcal{G} = (\mathbf{V}, \mathbf{E}) \), where \( \mathbf{V} \) represents variables and \( \mathbf{E} \) denotes directed edges encoding causal relationships, the goal is to ensure that the learned model \( f: \mathbf{X} \to Y \) adheres to the causal structure encoded in \( \mathcal{G} \)~\cite{berrevoets2024causal, zinati2024groundgan,afonja2024llm4grn,Scholkopfetal16}. 

\textbf{Invariant feature learning (IFL).} IFL relies on discovered \textit{implicit} or latent causal features and structures. The task of Invariant Feature Learning (IFL) is to identify features of the data \( \mathbf{X} \) that are predictive of the target \( Y \) across a range of environments \( \mathcal{E} \). From a causal perspective, the causal parents \( \text{Pa}(Y) \) are always predictive of \( Y \) under any interventional distribution~\cite{kaddour2022causal}. IFL can be achieved by regularizing the model or providing causal training data that is free of confounding. 

\textbf{Disentangled VAEs. } \
VAEs aim to decompose the data $\mathbf{X}$ into disentangled latent factors $\mathcal{Z}$ that correspond to distinct underlying generative causes, especially when trained with causal constraints or interventional data~\cite{burgess2018understanding, kim2021counterfactual}. This aligns with causal reasoning, as each latent dimension can be interpreted as an independent causal factor influencing the observed data. 


\textbf{Double Machine Learning (DML).} DML provides another causal approach by leveraging modern ML techniques for estimating high-dimensional nuisance parameters while preserving statistical guarantees in causal inference~\cite{chernozhukov2018double}. DML decomposes the estimation problem into two stages: (1) predicting confounders using ML models and (2) estimating the causal effect using residualized outcomes. 

\textbf{Causal Discovery.} Finally, ML can be leveraged for causal inference or to discover causal knowledge from observational data. For instance, methods for causal discovery use conditional independence, score, or noise asymmetry-based methods, to infer causal relationships, with notable examples including~\cite{spirtes2000causation,shimizu2006linear,janzing2010causal,PetMooJanSch11, hauser2012gies,le2016fast}. More recent methods leverage Deep Neural Networks (DNNs) in combination for causal discovery~\cite{nauta2019causal,bi2023large,  goudet2017causal}.

\textit{Summary and Use Cases.}

The methods presented above represent a spectrum of causal integration strategies in ML. CCML requires explicit or semi-explicit causal structures, offering high interpretability but often demanding strong prior knowledge. However, explicit causal guardrails are beneficial for high-stakes applications requiring clear accountability. In contrast, IFL, DML, and Disentangled VAEs operate with weaker structural assumptions and are well-suited for settings with limited causal annotations as well as exploratory tasks or scalable representation learning in unstructured data. Causal Discovery lies at the other end of the spectrum, aiming to uncover causal relationships from data alone. Combining methods—for example, using causal discovery to inform CCML also help balance rigor and practicality.

\section{Causality for Trustworthy Foundation Models} \label{sec:caualfoundation}

Foundation models (FMs), including state-of-the-art multimodal systems like LLMs and vision-language models, have demonstrated exceptional capabilities across diverse tasks~\cite{achiam2023gpt, team2023gemini, radford2023robust}. However, their reliability remains a concern due to issues like spurious correlations, hallucinations, and unequal representation. Trade-offs and causality in trustworthy foundation models remain underexplored as compared to traditional ML. In this section, we explore the potential for causality to improve fairness, explainability, privacy, and robustness in foundation models following a slightly different taxonomy than in the previous section due to their unique challenges.

\subsection{Dimensions of Trustworthy Foundation Models}
 In this section, we examine foundation model-specific trade-offs between key dimensions of trustworthy AI and illustrate how causal approaches can soften those tensions.

\textbf{Fairness vs. Accuracy.}
Causal frameworks support fairness in LLMs by identifying and mitigating pathways that lead to biased predictions~\cite{d011b9cc8bad2d29595f1ef5cff344abd2c5c5ab,c9db178549df34ea26994753e3560afac761f39e}. Techniques like counterfactual fairness ensure that sensitive attributes (e.g., gender, race) do not influence outcomes, while causal disentanglement separates spurious correlations from true predictive factors~\cite{zhou2023causal, 3a2e333502f5035f0f78318c7722ec1a7b7265bb}. Structural causal models (SCMs) further enable fairness-aware fine-tuning by isolating causal effects~\cite{d011b9cc8bad2d29595f1ef5cff344abd2c5c5ab}.

In addition, in generative models, careless fairness interventions can backfire—for instance, efforts to enhance diversity have led to historically inaccurate outputs, as seen in critiques of Google's Gemini~\cite{verge2024gemini}. Causal reasoning can help balance fairness and factuality by distinguishing meaningful diversity from implausible scenarios. It also offers tools to address mode collapse, preserving minority representation by capturing underrepresented causal patterns.

\textbf{Privacy vs. Attribution.} 
Causal approaches to privacy focus on identifying and blocking pathways involving personally identifiable information (PII) in LLMs. Causal obfuscation uses SCMs to remove only privacy-sensitive dependencies (e.g., names, locations), preserving essential predictive signals~\cite{chu2024causal}.

Attribution and memorization raise further concerns. Attribution determines whether specific data, such as an artist’s work, contributed to model training, while memorization complicates data removal. Causal auditing~\cite{sharkey2024causal} offers a principled alternative to correlation-based methods by disentangling direct data influences from incidental stylistic similarities, enabling more accurate and accountable attribution.



\textbf{Robustness vs. Accuracy.} Causal frameworks address robustness by training models to rely on invariant causal relationships while penalizing reliance on dataset-specific spurious features~\cite{9f597bdbd47d38e3920d2ab9020b91a558cdf026}. For instance, instead of associating "doctor" with "male," causal invariance enforces reliance on task-relevant features like medical terminology~\cite{zhou2023causal}. Causal regularization further discourages attention to non-causal patterns during inference, achieving better accuracy and robustness.

\textbf{Explainability vs. Capability.} 
Despite their impressive capabilities, foundation models often lack interpretability, making their reasoning difficult to explain. Causal models address this by quantifying how input features influence outputs and modeling interactions across the model’s components (e.g., embeddings, attention, logits)\cite{bagheri2024c2p}. This enables step-by-step explanations of the model’s decisions. Closely related is mechanistic interpretability, which analyzes model architecture, weights, and activations \cite{conmy2023towards}. Interventional methods enhance this by uncovering cause-effect relationships within neural circuits~\cite{palit2023towards, parekh2024concept}. Application of formal causal theory and methods to mechanistic interpretability is an emerging research direction that shows promise for deeper, more robust insights~\cite{geiger2024causal, stolfo2023mechanistic, sharkey2025open, sharkey2024causal}.

\subsection{Integrating Causality in Foundation Models} 
\label{sec:integrasting}
This section delves into practical applications of causality in FMs across three key stages: pre-training, post-training, and auditing. We conclude with a discussion of the practical advantages and limitations of the proposed approaches.



\textbf{Causal Representation Learning.}
By disentangling causal factors from non-causal ones, models can learn representations that separate meaningful causal features from irrelevant associations. Techniques such as causal embedding methods~\cite{rajendran2024learning, jiang2024origins}, use training data annotated with causal labels.
This has been shown to reduce reliance on spurious correlations, such as gender-biased occupational associations~\cite{zhou2023causal}. Fine-tuning models on embeddings pre-trained with debiased token representations has shown promise for causal learning~\cite{kaneko2021debiasing, guo2022auto,he2022mabel, wang2023causal}.

\textbf{Entity interventions.}
SCMs can be used to determine interventions on specific entities (e.g., replacing ``Joe Biden" with ``ENTITY-A") during pre-training~\cite{wang2023causal}, thus reducing entity-based spurious associations while preserving causal relationships in the data.




\textbf{Counterfactual Data Augmentation.} 
Synthetic datasets with explicit causal structures or counterfactual examples, that introduce scenarios where causal relationships differ from spurious correlations, help models learn true causal dependencies instead of misleading patterns~\cite{webster2020measuring, chen2022disco}. 
Synthetic causal data can also be used for fine-tuning the models, similar to pre-training, but with better sample size efficiency. Frameworks like DISCO~\cite{chen2022disco} generate diverse counterfactuals during fine-tuning to enhance OOD generalization for downstream tasks. 

\textbf{SCM-based Methods.}
Causally Fair Language Models (CFL)~\cite{d011b9cc8bad2d29595f1ef5cff344abd2c5c5ab} use SCM-based regularization to detoxify outputs or enforce demographically neutral generation during post-training. ~\cite{wang2020identifying} use causal reasoning to separate genuine from spurious correlations by computing controlled direct effects, ensuring robust performance. 

\textbf{Causal RLHF Alignement.} 
RLHF can be adapted to include causal interventions, allowing feedback to act as instrumental variables that correct biased model behavior. Causality-Aware Alignment~\cite{xia-etal-2024-aligning} incorporates causal interventions to reduce demographic stereotypes during fine-tuning with alignment objectives. Extending RLHF with causal alignment to support dynamic, context-sensitive interventions could help address biases that evolve. Integrating causal reasoning into the reward model's decision-making process, by critiquing the output of LLM using a reward model or a mixture of reward models that control for specific confounders or spurious correlations, can potentially improve the downstream reasoning abilities, potentially mitigating hallucinations.

\textbf{Latent Representation Discovery.}
Recent work in mechanistic interpretability has employed sparse autoencoders to extract interpretable latent representations aligned with internal circuit features~\cite{kissane2024interpreting, conmy2023towards}. These methods enable the identification of functional substructures within models by isolating meaningful directions in activation space. 

\textbf{Targeted Interrogation Methods.}
Techniques based on interventions, such as ablation, activation patching, or causal scrubbing, allow us to move beyond passive observation by directly testing the functional roles of model components~\cite{brinkmann2024mechanistic,syed2023attribution}. These methods provide a powerful framework for identifying how specific neurons or pathways contribute to behavior, enabling rigorous, theory-driven interpretability.


\textbf{Causal Discovery.} The recent advancements in large language models (LLMs) have inspired their use in causal discovery~\citep{kiciman2023causal, kasetty2024evaluating, vashishtha2023causal, ai4science2023impact, abdulaal2023causal,khatibi2024alcm}. 
Most of the above methods involve the refinement of the statistically inferred causal graph by LLM. However, emerging research shows that LLMs can be useful in constructing full causal graphs based on parametric knowledge of scientific literature in diverse domains~\citep{sheth2024hypothesizing,afonja2024llm4grn}. 



\textit{Summary and Use Cases.}

These methods span both causal inference and integration. Causal data augmentation, representation learning, and entity interventions are methods used in pretraining, where broad structural priors can be encoded. In contrast, causal fine-tuning, alignment, and auditing are post-training methods requiring less causally annotated data, but more explicit goals and causal structures. Latent representation discovery and targeted interrogation methods stand out as a retrospective techniques that can help to assess fairness, privacy, and robustness without modifying the training pipeline. Causal discovery can support both pre- and post-training by informing model design or evaluating learned representations. Selecting among these approaches depends on the availability of causal knowledge, the interpretability needs, and the constraints of the training pipeline.



\section{Challenges and Opportunities}\label{sec:challenges}
Despite the advantages of a causal approach, practical applications pose challenges, including reliance on strong causal assumptions and limited availability of \textit{a priori} causal knowledge. FMs bring further complications due to their complexity and scale. We outline key obstacles in integrating causality into ML and FMs and suggest strategies to overcome them. 

\subsection{General Challenges in Applying Causality} 

\textbf{Availability of Causal Knowledge.}
A major challenge in causal ML is the limited availability of causal knowledge, particularly in the form of DAGs. Expert-constructed DAGs may suffer from subjectivity and scalability issues, while ML-based causal discovery is constrained by identifiability assumptions and noise sensitivity. However, recent hybrid approaches combining classical causal discovery with LLM-based reasoning offer promising solutions. In addition, approximate causal interventions are possible without fully specified model or with partial causal graph~\cite{russell2017worlds,ehyaeibridging,zuo2022counterfactual}.

\textbf{Causal Transportability.} Scientific knowledge often lacks direct applicability across different populations, making causal transportability essential. Pearl and Bareinboim's DAG-based framework adjusts causal knowledge for new settings using targeted data collection~\cite{bareinboim2014transportability, pearl2011transportability}. Building on this, \cite{binkyte2024babe} propose an expectation-maximization (EM) approach to adapt causal knowledge for target demographic applications.



\textbf{Potentially Unresolvable Tensions.}
Not all tensions in trustworthy AI can always be fully resolved. For instance, stronger privacy protections often reduce model utility~\cite{dwork2014algorithmic, bassily2014private}.  We also acknowledge that some fairness conflicts stem from deeper normative or value-based disagreements, for example, a causal relationship may exist, but relying on it in decision-making may still be viewed as unfair from an ethical or legal standpoint. In such cases, causality does not offer a definitive solution but provides a framework to surface these assumptions transparently, enabling more informed discussions that go beyond technical considerations.


\subsection{Challenges in Causal Foundation Models} 
 \textbf{Concept Superposition.} One foundation model-specific challenge is concept superposition, particularly in LLMs, where multiple meanings are entangled within a single representation, complicating causal reasoning~\cite{elhage2022toy}. Vision models exhibit this issue to a lesser extent due to their structured data formats. Being aware of superposition is imporant for effectively integrating causality.

 \textbf{Lack of High-quality Causal Data.} Training foundation models with causal reasoning requires datasets annotated with explicit causal structures or interventional data, which are scarce and expensive to produce. Scalable methods for generating synthetic causal datasets show a promising direction~\cite{webster2020measuring, chen2022disco}. Alternatively, focusing on post-training methods allows causal interventions in a more data-efficient way.

\textbf{Computational Complexity.} Additionally, the computational complexity of integrating causal reasoning into foundation models poses a significant challenge. However, low-rank adaptation methods such as LoRA can be employed to reduce the number of learnable parameters, making causal integration more efficient without compromising performance~\cite{hu2021lora}.

\section{Alternative View}
\label{sec:Alternative}


Symbolic AI leverages prior knowledge through logic rules, ontologies, and formal representations, supporting structured reasoning, explainability, and generalization from small data~\cite{diaz2022explainable}. It can also help navigate trade-offs, for  example, between accuracy and interoperability, by making the reasoning process more transparent and controllable. While both symbolic AI and causal reasoning rely on prior assumptions, causality uniquely enables reasoning about interventions and counterfactuals. Unlike symbolic systems that model static relationships, causal models capture how changes in one variable affect others - an essential feature in domains like fairness and policy evaluation.

Another perspective suggests that causal properties may emerge spontaneously as models scale and are trained on increasingly diverse data. Recent theoretical work by~\cite{richens2024} suggests that robust agents implicitly learn causal world models if exposed to sufficiently diverse environments. However, this emergence is neither guaranteed nor controllable. In contrast, explicitly integrating existing scientific causal knowledge into machine learning provides a more reliable, interpretable, and resource-efficient pathway to trustworthy AI.


\section{Conclusion and Call for Action}\label{sec:conclusion}
Causal models offer a principled approach to trustworthy AI by prioritizing relationships that are causally justified and invariant across contexts. This approach reduces tensions between competing objectives and can enhance multiple dimensions—privacy, accuracy, fairness, explainability, and robustness—simultaneously, creating models that are not only ethically sound but also practically effective.

To further advance trustworthy ML foundation models, we emphasize the need for the following actions:

\textit{Incorporate Trade-off Awareness in Model Design}: Ensure that foundation models are developed with explicit consideration of trade-offs between key trustworthy AI dimensions—fairness, privacy, robustness, explainability, and accuracy. 

\textit{Leverage Causality to Resolve or Soften Trade-offs}: Where possible, integrate causal reasoning to disentangle competing objectives and mitigate conflicts. 

\textit{Develop Scalable Methods for Causal Data Integration}: Encourage the development of algorithms and pipelines to integrate causal knowledge into foundation models at scale. 

\textit{Create and Share High-Quality Causal Datasets}: Foster initiatives to curate, annotate, and share datasets with explicit causal annotations or interventional information.

\textit{Advance Causal Discovery Techniques}: Invest in research to improve causal discovery algorithms. 
Hybrid approaches combining classical methods with LLM-based contextual reasoning show a promising direction.

\textit{Benchmark and Evaluate Causal Models}: Establish evaluation frameworks that assess the ability of causal models to balance trade-offs effectively and provide transparent justifications for their decisions in high-stakes domains.

All these advancements are crucial for expanding the application of causality in ML and foundation models, paving the way for more balanced and trustworthy AI solutions.

\bibliographystyle{unsrt}
\bibliography{references.bib}

\end{document}